\documentclass[sigconf]{acmart}

\AtBeginDocument{%
  }

\usepackage{multirow}
\usepackage{multicol}

\usepackage{amssymb}
\usepackage{graphicx}
\usepackage{enumitem}
\usepackage{url}
\usepackage{balance} 
\usepackage{pdfpages}
\usepackage{xstring}
\usepackage{xspace}
\usepackage{subcaption}
\usepackage{array}
\usepackage{balance} 
\usepackage{color}
\usepackage{multirow}
\usepackage{booktabs}
\usepackage{xcolor,colortbl}

\usepackage{xurl}
\usepackage{seqsplit}

\setlist[itemize,1]{leftmargin=\dimexpr 26pt-2mm}
\definecolor{MapAgentBG}{RGB}{235,245,255}   
\definecolor{MapAgentRed}{RGB}{180,30,30}

\usepackage[english]{babel}
\usepackage{microtype}

\copyrightyear{2026}
\acmYear{2026}
\setcopyright{cc}
\setcctype{by}
\acmConference[KDD '26]{Proceedings of the 32nd ACM SIGKDD Conference on Knowledge Discovery and Data Mining V.2}{August 09--13, 2026}{Jeju Island, Republic of Korea}
\acmBooktitle{Proceedings of the 32nd ACM SIGKDD Conference on Knowledge Discovery and Data Mining V.2 (KDD '26), August 09--13, 2026, Jeju Island, Republic of Korea}
\acmDOI{10.1145/3770855.3818443}
\acmISBN{979-8-4007-2259-2/2026/08}

\begin{document}

\title{MapAgent: An Industrial-Grade Agentic Framework for City-scale Lane-level Map Generation}

\author{Deguo Xia}
\authornote{These authors contributed equally to this work.}
\email{xdg23@mails.tsinghua.edu.cn}
\affiliation{%
  \institution{Tsinghua University}
  \institution{Baidu}
  \state{Beijing}
  \country{China}}
\orcid{0000-0003-3366-2230}

\author{Zihan Li}
\authornotemark[1]
\orcid{0009-0009-4347-2239}
\email{mc45085@um.edu.mo}
\affiliation{%
  \institution{University of Macau}
  \state{Macao}
  \country{China}}

\author{Haochen Zhao}
\authornotemark[1]
\email{zhaohaochen@iie.ac.cn}
\orcid{0009-0006-8593-6079}
\affiliation{%
  \institution{Institute of Information Engineering, Chinese Academy of Sciences}
  \city{Beijing}
  \country{China}
  }

\author{Dong Xie}
\authornotemark[1]
\email{xiedong04@baidu.com}
\orcid{0009-0003-6358-0823}
\affiliation{%
  \institution{Baidu}
  \state{Beijing}
  \country{China}}
  
\author{Yuyao Kong}
\email{kongyuyao@iie.ac.cn}
\orcid{0009-0005-4136-7356}
\affiliation{%
  \institution{Institute of Information Engineering, Chinese Academy of Sciences}
  \city{Beijing}
  \country{China}
  }

\author{Xiyan Liu}
\email{liuxiyan@baidu.com}
\orcid{0000-0002-0102-9636}
\affiliation{%
  \institution{Baidu}
  \state{Beijing}
  \country{China}}
  
\author{Jizhou Huang}
\authornote{Corresponding authors.}
\email{huangjizhou01@baidu.com}
\orcid{0000-0003-1022-0309}
\affiliation{%
  \institution{Baidu}
  \state{Beijing}
  \country{China}}

\author{Mengmeng Yang}
\authornotemark[2]
\email{yangmm_qh@tsinghua.edu.cn}
\orcid{0000-0002-3294-6437}
\affiliation{%
  \department{School of Vehicle and Mobility}
  \department{State Key Laboratory of Intelligent Green Vehicle and Mobility}
  \institution{Tsinghua University}
  \state{Beijing}
  \country{China}}
  
\author{Diange Yang}
\email{ydg@mail.tsinghua.edu.cn}
\orcid{0000-0002-0074-2448}
\affiliation{%
  \department{School of Vehicle and Mobility}
  \department{State Key Laboratory of Intelligent Green Vehicle and Mobility}
  \institution{Tsinghua University}
  \state{Beijing}
  \country{China}}

\renewcommand{\shortauthors}{Deguo Xia et al.}

\begin{abstract}
Lane-level maps are critical infrastructure for autonomous driving and lane-level navigation, yet constructing and maintaining standardized lane networks for hundreds of cities remains highly labor-intensive. Recent end-to-end vectorized mapping methods can predict lane geometry and topology directly from sensor data, but they typically treat mapping specifications and traffic regulations as implicit, dataset-dependent supervision. Moreover, in complex scenes (e.g., worn or missing markings and occlusions), correct lane configurations are often under-determined by visual evidence alone, making specification violations a major source of human post-editing.
We propose MapAgent, an industrial-grade agentic architecture that augments a vectorization backbone for specification-compliant lane-map production. Rather than merely adding an agent loop to map prediction, MapAgent couples backbone perception with explicit specification verification, constraint-aware reasoning, and deterministic map editing under a bounded, verification-driven Judge--Planner--Worker loop. A vision--language Judge diagnoses errors by jointly inspecting visual evidence and draft vectors, while a tool-calling Planner generates minimal corrective edits with post-edit re-validation. 
To remain scalable for city-scale production, MapAgent is selectively triggered only on tiles with low backbone confidence, adding modest overhead while preserving throughput.
Experiments on real-world datasets show consistent gains over strong production baselines, especially in complex and long-tail scenarios. 
Additionally, MapAgent has been integrated into Baidu Maps, supporting lane-level map generation for over $360$ cities nationwide and elevating the overall production automation to over $95\%$, demonstrating MapAgent’s practicality and effectiveness for large-scale lane-level map generation.

\end{abstract}

\ccsdesc[300]{Applied computing~Transportation}

\keywords{Map Generation; Map Agent; Vision-Language Models}

\maketitle

\section{Introduction}
Lane-level maps have become core infrastructure for autonomous driving, advanced driver assistance, and lane-level navigation. They provide centimeter-level priors on road geometry, lane topology, and traffic control, enabling robust planning and decision-making beyond the sensing horizon. However, constructing and maintaining lane-level maps for hundreds of cities at nationwide scale remains extremely expensive. 
Traditional pipelines rely heavily on trained experts to interpret mapping rules and standards and to conduct labor-intensive annotation and updating, which limits nationwide coverage and update cycles ~\cite{li2022hdmapnet,xia2024dumapnet,xia2025ldmapnet}. 
Recent end-to-end HD map learning has pushed automation from research to industry: methods such as HDMapNet~\cite{li2022hdmapnet}, VectorMapNet~\cite{liu2023vectormapnet}, MapTR~\cite{liao2022maptr} and MapTRv2~\cite{liao2023maptrv2} convert multi-sensor inputs into bird's-eye-view (BEV) features and directly decode vectorized polylines or topology, replacing much of the manual mapping pipeline while achieving strong performance. 
To meet nationwide requirements on scale, efficiency, and quality, in our prior work, we developed DuMapNet~\cite{xia2024dumapnet}, an industrial vectorization system for city-scale lane-level map generation in Baidu Maps (deployed since 2023), and subsequently developed LDMapNet-U~\cite{xia2025ldmapnet} for map updating conditioned on historical maps (deployed since April 2024); together, they support over 360 cities and substantially reduce production cost and update latency.

\begin{figure}[h]
  \centering
  \includegraphics[width=0.48\textwidth]{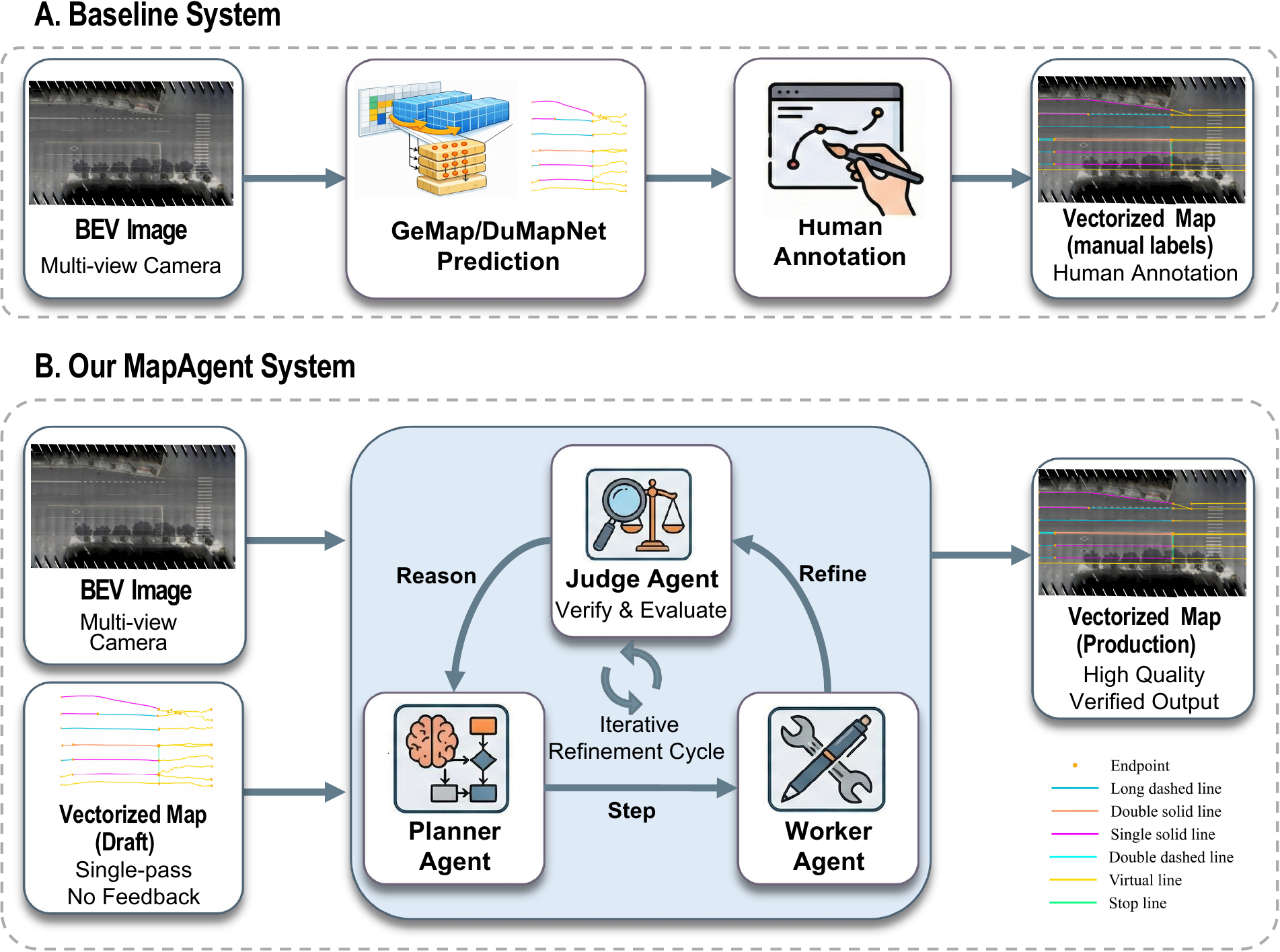}
    \caption{\textbf{Framework Comparison.}
    \textbf{Top:} A one-pass pipeline vectorizes lanes from BEV images and relies on human post-editing.
    \textbf{Bottom:} MapAgent performs agent-based refinement with a Judge--Planner--Worker loop to automatically verify errors, apply corrections, and produce a high-quality map.}
  \label{fig:compare}
\end{figure}

\begin{figure*}[t]
  \centering
  \includegraphics[width=0.95\linewidth]
  {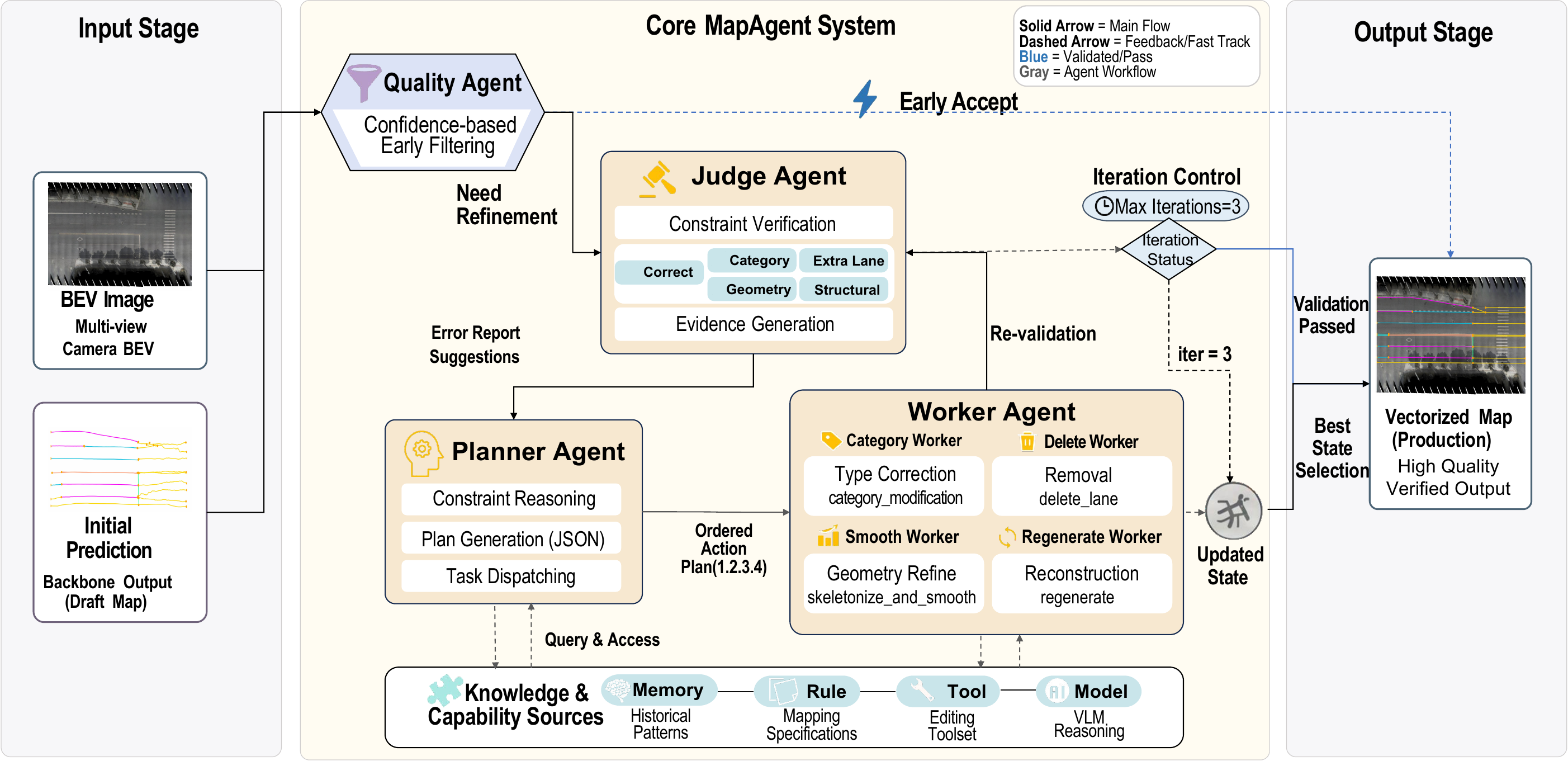}
  \caption{Overall MapAgent system. Given a BEV image and a draft vector map from a backbone, a \textbf{Quality Agent} first performs confidence-based early filtering. High-confidence tiles are directly accepted, while the rest enter a bounded \textbf{Judge--Planner--Worker} refinement loop. The Judge detects rule violations and produces structured reports, the Planner generates an ordered correction plan under constraints, and the Worker applies deterministic edits such as removal, type correction, smoothing, or regeneration. Each iteration is re-validated, and after at most three rounds, the system outputs the validated refined map or the best intermediate result.}
  \label{fig:framework}
\end{figure*}

Despite these advances, a crucial gap remains between end-to-end vectorization and fully automated, specification-compliant lane-level map production. 
Existing systems mainly learn what is visible---lane boundaries, crosswalks, stop lines, and local topology---from supervised labels~\cite{li2022hdmapnet,liu2023vectormapnet,liao2022maptr}, whereas industrial maps must also satisfy cartographic standards and traffic regulations, requiring consistent organization of lane groups, geometry, attributes, and topology.
In long-tail real-world scenes (e.g., degraded/missing markings, wide unstructured pavements, adverse lighting, and occlusions), the lane configuration is frequently under-determined by visual evidence alone and instead depends on specification- and rule-based priors, leading supervised models to exhibit geometric artifacts and semantic misclassifications that still require substantial human post-editing, even in city-scale deployments such as DuMapNet and LDMapNet-U.
As a result, commercial pipelines still depend on expert editors who apply codified specifications via interactive tools to repair topology and ensure compliance---capabilities that current one-pass models lack, since they neither explicitly represent such knowledge nor offer a principled way to reason about per-scene ambiguities or to decide when and how to invoke geometric editing tools.

Motivated by this gap, we propose a shift from one-pass end-to-end vectorization to an industrial-grade paradigm of \emph{agent-based refinement on top of a frozen backbone}, where the backbone generates drafts and an agentic layer enforces specifications via verification and deterministic edits.
In this setting, a BEV vectorization backbone remains essential for scalability and visual performance, but it is treated as a draft generator rather than the sole component responsible for satisfying cartographic and regulatory constraints. We therefore refine frozen backbone outputs with a controllable workflow that combines grounded diagnosis, specification-aware verification, and deterministic tool-based edits. To realize this, we introduce MapAgent, a refinement-on-top-of-backbone framework that selectively processes only hard tiles through a bounded iterative loop. A lightweight Quality Agent performs early acceptance using backbone confidence and cheap consistency checks; for the remaining tiles, MapAgent runs a Judge--Planner--Worker loop where a vision--language Judge produces structured error evidence by verifying geometry, topology, and specification compliance, a Planner converts evidence into a tool-grounded edit plan under capability constraints, and deterministic Workers execute edits with re-validation until success or budget exhaustion. By coupling end-to-end perception with explicit verification and rule-grounded editing under strict safety constraints (closed tools, bounded budget, best-state fallback), MapAgent enables scalable, production-quality lane-level maps aligned with domain standards.

Concretely, this paper makes the following key contributions to both the research and industrial communities:

\begin{itemize}
\item \textbf{Potential impact:} We introduce MapAgent as an industrial-grade agentic refinement layer for city-scale lane-level map generation and updating. MapAgent has been integrated into Baidu Maps, supporting lane-level map generation and updating for over $360$ cities nationwide, and elevating the overall production automation to over $95\%$.
\item \textbf{Novelty:} MapAgent introduces an industrial-grade agentic paradigm for lane-level mapping, designed for specification-compliant production. By coupling a frozen BEV vectorization backbone with a bounded, verification-driven loop (Quality Agent + Judge--Planner--Worker), it enforces hard validity checks for geometric, topological, and specification constraints.
\item \textbf{Technical quality:}  Extensive experiments on large-scale real-world datasets demonstrate consistent gains over strong baselines, especially in complex and long-tail scenarios, while substantially reducing manual post-editing effort. The successful production integration of MapAgent further validates its robustness and scalability for city-scale deployment.
\end{itemize}

\section{MapAgent Framework}
\label{sec:framework}

Compared to conventional one-pass pipelines (\emph{BEV observation} $\rightarrow$ \emph{vector decoding}  $\rightarrow$ \emph{human annotation} $\rightarrow$ \emph{final map}),
MapAgent reconceptualizes lane-level map generation as a controllable refinement process: the backbone output is treated as a mutable
map state rather than a final result, as shown in Figure \ref{fig:compare}. The system verifies the draft map using a structured Judge, plans minimal corrective actions,
and executes deterministic edits via tools, making the pipeline efficient, interpretable, and production-ready. In the following subsections, we first formalize MapAgent as a constrained iterative refinement process. We then introduce the overall system architecture and refinement workflow, followed by detailed descriptions of the Judge Agent for constraint verification, the Planner Agent for tool-based plan generation, and the deterministic Worker agents for map editing.

\subsection{Problem Formulation}
\label{sec:problem_formulation}

Following prior works on lane-level map generation and updating~\cite{xia2024dumapnet,xia2025ldmapnet}, the general task is to convert a BEV observation into a standardized vectorized lane map. Concretely, given a BEV observation $I$ collected from vehicle-mounted sensors, a map generation or updating system aims to predict a structured vectorized map $V$, where each map element is represented in a unified vectorized form with its geometry and attributes. 
In this paper, we formulate MapAgent as a constrained map refinement problem built on top of a frozen BEV vectorization backbone.
Given a BEV observation $I$ and an initial draft map $V_0$ produced by the backbone, MapAgent iteratively edits $V_t$ into a specification-compliant map.

At refinement step $t$, we treat $s_t = (I, V_t) \in\mathcal{S}$ as the environment state.
Let $\mathcal{L}$ denote the set of low confidence lane instances selected by the Quality Agent for refinement.

The Judge $\mathcal{J}_\phi$ produces a \emph{lane-wise} structured diagnosis for each $l\in\mathcal{L}$,
\begin{equation}
\mathsf{S}_{\mathrm{judge}}(l) = \mathcal{J}_\phi\big(I, V_t, l\big),
\end{equation}
and we aggregate them into a step-level diagnosis
\begin{equation}
E_t = \{\mathsf{S}_{\mathrm{judge}}(l)\mid l\in\mathcal{L}\}.
\end{equation}
Conditioned on $(V_t,E_t)$, the Planner is implemented as a rule-based module $g(\cdot)$ that generates a tool-grounded refinement plan:
\begin{equation}
\mathcal{P}_t=g(V_t,E_t),\quad \mathcal{P}_t\in\mathcal{A}^* .
\end{equation}
Concretely, $\mathcal{P}_t$ is an ordered sequence of actions
\begin{equation}
\mathcal{P}_t = (a_{t,1}, a_{t,2}, \dots, a_{t,K_t}), \qquad a_{t,k} \in \mathcal{A}.
\label{eq:p_action}
\end{equation}
Each action follows a fixed schema (consistent with Worker tools), e.g.,
\(a_{t,k} = (\texttt{tool}, \texttt{lane\_id}, \texttt{params})\), and the plan is executed deterministically to update the map via
\begin{equation}
V_{t+1} = \mathcal{T}(V_t, \mathcal{P}_t).
\end{equation}
All edits must satisfy an immutable feasibility gate $\Omega$ (geometric and topological validity, mapping specifications); edits 
that lead to $\Omega(V_{t+1})=0$ are rejected by design (see App.~\ref{apsec:omega}).
Overall, MapAgent is driven by a learned Judge $\mathcal{J}_\phi$ that produces structured diagnoses $E_t$ (trained via SFT and RL; see Sec.~\ref{sec:judge}), and a rule-based Planner $g(\cdot)$ that generates minimal valid plans under $\Omega$ using a closed action set defined by Worker tools.

\begin{figure*}[t]
  \centering
  \includegraphics[height=8.5cm, width=17cm]{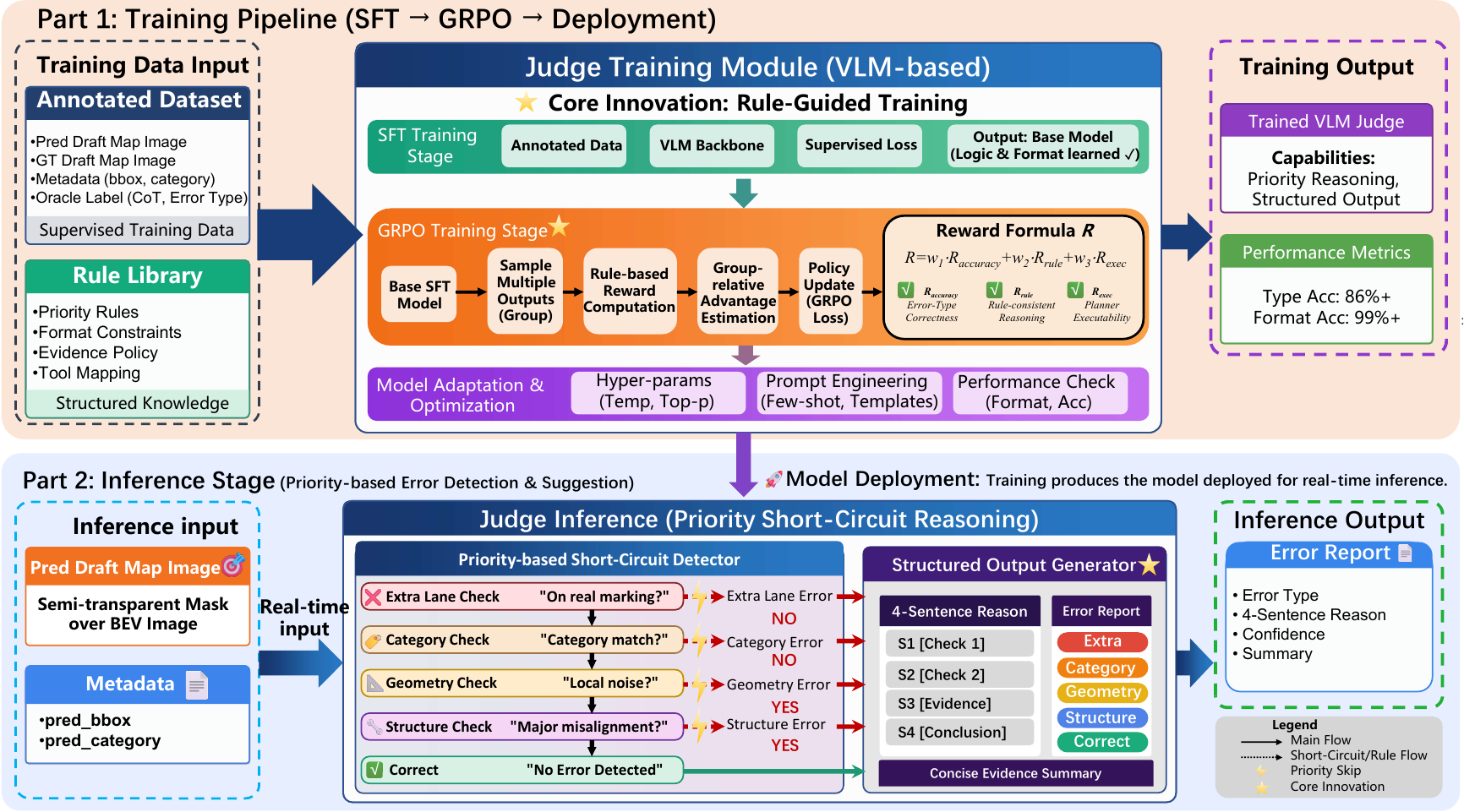}
  \caption{Architecture of the VLM-based Judge Agent. The framework consists of two phases. \textbf{Top:} A rule-guided training pipeline that aligns the VLM via SFT followed by GRPO, using a composite reward to enforce rule compliance, reasoning consistency, and executability. The training leverages annotated data and structured rule libraries to learn priority reasoning and structured outputs. \textbf{Bottom:} At inference, the Judge takes a BEV image and a draft map, applies a priority-based short-circuit mechanism to check errors in order, and generates structured, evidence-backed reports. The output includes error type, concise reasoning, and confidence, which are used by the Planner for downstream refinement.}

  \label{fig:judge_arch}
\end{figure*}

\subsection{Overall Architecture}
\label{framework}

As illustrated in Figure~\ref{fig:framework}, MapAgent is an agent-based refinement framework on top of a frozen backbone. It builds upon city-scale BEV vectorization backbones to retain strong perception performance and scalability, while introducing an explicit agentic verification-and-editing loop that enforces lane-level specifications.
The Quality Agent applies an early-acceptance fast track to tiles whose confidence score exceeds a fixed threshold $\delta=0.7$, while only tiles with confidence below $\delta$ are forwarded to a bounded refinement loop.
This selective routing focuses computation on difficult regions while preserving the throughput of the backbone. 
In practice, most tiles are directly accepted, and only a minority of hard tiles require iterative refinement.
The separation between fast-track acceptance and iterative correction also simplifies system integration, as the backbone remains unchanged.

For each forwarded tile, MapAgent runs an iterative Judge--Planner--Worker workflow with re-validation.
We explicitly bound the outer refinement budget by a small constant, three rounds in our system, to ensure predictable latency and prevent over-editing.
This bounded design provides stable runtime behavior and avoids cascading modifications.
At each round, the system evaluates the current map state before applying the next plan, ensuring that only validated edits are retained.
If the refinement converges earlier, the loop terminates immediately.
When the budget is exhausted without further admissible improvement, the system outputs the best validated map state.

\subsection{Judge Agent: Constraint Verification and Evidence Generation}
\label{sec:judge}
As shown in Figure~\ref{fig:judge_arch}, the Judge Agent conducts geometry and topology verification, enforces specification compliance, and produces structured evidence for downstream refinement.

For each inspected lane \(l\), the Judge produces a structured, scalar-sized diagnostic entry
\begin{equation}
S_{\mathrm{judge}}(l) \;=\; \big(l,\ \hat y_l,\, \hat c_l, e\big),  
\end{equation}
where \(\hat y_l \in \mathcal{C}\) is the predicted error category drawn from the predefined error set \(\mathcal{C}\), \(\hat c_l\in[0,1]\) is the associated confidence, and $e$ is a concise evidence summary. 
The taxonomy \(\mathcal{C}\) is derived from large-scale historical post-editing logs; newly observed anomalies are reviewed through a human-in-the-loop process, clustered, and used to update mapping specifications and subsequent Judge retraining data.
This fixed schema \(S_{\mathrm{judge}}(l)\) is the sole interface exposed to the Planner: internal artifacts (e.g., chain-of-thought traces) are retained for training and debugging only and are not directly consumed by downstream modules. 
Applying the Judge to all $l\in\mathcal{L}$ yields lane-wise diagnoses $\{S_{\text{judge}}(l)\}$,
which are aggregated as $E_t$ and passed to the Planner at round $t$.

To create supervised training instances, we sample individual lanes from the backbone draft map and construct model inputs by overlaying the target lane $l$ onto the BEV observation $I$  as a semi-transparent mask to focus attention on local evidence. 
Each instance is paired with a ground-truth error label $y_l\in\mathcal{C}$ and, optionally, an explanatory reference trace used for SFT. 
We denote a single supervised trajectory for lane $l$ as ${\tau}_l=(\mathcal{R}_1,\dots,\mathcal{R}_M,y_l)$, where $\mathcal{R}_m$ denotes the $m$-th intermediate reasoning rationale, and the supervised corpus as $\mathcal{D}=\{{\tau}_l\}$. 
Note that the lane-masking strategy is a training-time input construction intended to focus attention on local evidence while preserving the global context through the unmasked portions of \(I\).
Thus, the Judge is lane-centered but not lane-isolated: it still conditions on the full BEV observation and current map context, allowing it to reason about consistency with adjacent lanes.

The Judge Agent is implemented as an autoregressive multimodal policy $\pi_\phi$ that conditions on the BEV $I$, the current map context $V_t$, the selected lane $l$, and previously generated tokens to produce a short reasoning sequence followed by the final structured prediction $S_{\mathrm{judge}}(l)$. 

We first perform supervised fine-tuning on $\mathcal{D}$ by maximizing the likelihood of the reference traces and final labels. To maintain stability and avoid catastrophic changes to the pretrained backbone, only lightweight adapter parameters are updated during SFT while the backbone weights remain frozen.

After SFT, we further apply GRPO to align the Judge with the downstream Planner--Worker objective. 
Given a lane-level input $x=(I,V_t,l)$, the Judge policy generates an output $y$ consisting of a concise reasoning trace, a final error type, a confidence score, and an evidence summary. 
For each input $x$, GRPO samples a group of candidate outputs from the old policy,
\begin{equation}
\{y_1,\ldots,y_G\}\sim \pi_{\phi_{\mathrm{old}}}(\cdot|x),
\end{equation}
and computes a scalar reward $R_i$ for each candidate $y_i$. 
Unlike PPO, GRPO does not require a separate value model; instead, it estimates the advantage by normalizing rewards within the sampled group:
\begin{equation}
A_i=
\frac{
R_i-\mathrm{mean}(\{R_j\}_{j=1}^{G})
}{
\mathrm{std}(\{R_j\}_{j=1}^{G})+\epsilon
}.
\end{equation}
This group-relative normalization reduces memory cost during VLM fine-tuning and naturally fits our setting, where multiple candidate diagnoses for the same lane can be directly compared.

We optimize the clipped surrogate
\begin{equation}
\begin{aligned}
L_i(\phi) &=
\min\left(
\rho_i A_i,\,
\bar{\rho}_i A_i
\right), \\
\rho_i &=
\frac{\pi_\phi(y_i|x)}
{\pi_{\phi_{\mathrm{old}}}(y_i|x)}, \\
\bar{\rho}_i &=
\mathrm{clip}
\left(
\rho_i,
1-\epsilon_{\mathrm{clip}},
1+\epsilon_{\mathrm{clip}}
\right).
\end{aligned}
\label{eq:grpo-surrogate}
\end{equation}

The GRPO objective is then defined as
\begin{equation}
\begin{aligned}
\mathcal{J}_{\mathrm{GRPO}}(\phi)
=
\mathbb{E}_{x,\{y_i\}_{i=1}^{G}}
\left[
\frac{1}{G}
\sum_{i=1}^{G}
L_i(\phi)
-
\beta
D_{\mathrm{KL}}
\left(
\pi_\phi(\cdot|x)
\|
\pi_{\mathrm{ref}}(\cdot|x)
\right)
\right],
\end{aligned}
\label{eq:grpo-objective}
\end{equation}
where $\pi_{\mathrm{ref}}$ is the frozen SFT policy. The KL penalty prevents the updated Judge from drifting away from the structured output behavior learned during SFT.

For each candidate output, the reward is computed as
\begin{equation}
R_i
=
R_{\mathrm{acc}}(y_i)
+
0.5 R_{\mathrm{rule}}(y_i)
+
R_{\mathrm{exec}}(y_i).
\label{eq:grpo-reward}
\end{equation}
We compute the terms in an executability-first order. 
$R_{\mathrm{exec}}$ acts as a hard gate: if the response cannot be parsed as JSON, the reward computation stops with $R_{\mathrm{exec}}=-2.0$; if the JSON misses required fields or the evidence mentions lane identifiers outside the valid metadata set, we also assign a $-2.0$ executability penalty. 
A valid schema with no hallucinated lane identifier receives a small format reward $R_{\mathrm{exec}}=0.2$. 
$R_{\mathrm{acc}}$ is $+1.0$ when the predicted \texttt{error\_type} matches the oracle label and $-1.0$ otherwise. 
$R_{\mathrm{rule}}$ starts from zero and checks the reasoning trace: it subtracts $0.5$ if the trace does not contain exactly four sentences, subtracts $0.5$ if it mentions any lower-priority error type than the final prediction, and subtracts $0.3$ if required higher-priority exclusions are missing. 
If none of these rule violations occurs, $R_{\mathrm{rule}}=0.5$. 
During GRPO training, we update only the LoRA adapter parameters and keep the pretrained VLM backbone frozen, preserving visual grounding while improving Judge reliability on hard cases under industrial map-production constraints.

We do not train the full Judge–Planner–Worker stack end-to-end. This is intentional: the Planner encodes immutable mapping specifications and the Workers are deterministic executors, so exposing them to unconstrained policy optimization would weaken safety guarantees. The learned component is therefore restricted to the Judge, while Planner/Worker behavior remains auditable and version-controlled.

\subsection{Planner Agent: Tool-Grounded Plan Generation}
\label{sec:planner}

The Planner Agent is a decision-making module that converts lane-level quality
assessments produced by the Judge Agent into executable refinement plans under strict mapping constraints.

At refinement step $t$, the Planner receives the structured Judge outputs for candidate lane lines and consumes the pair $(V_t,E_t)$ to generate the corrective plan $\mathcal{P}_t$.
When required for consistency, the Planner may additionally query the historical map state \(V_{t-1}\) to reason about past modifications. 
This query acts as an anti-oscillation filter and is triggered only when the same lane is edited in consecutive iterations; in production logs, this occurs in less than $10\%$ of refinement cases. 
No perceptual features beyond these structured inputs are available to the Planner.

The Planner outputs a structured refinement plan $\mathcal{P}_t$ following a fixed schema. 
A plan consists of an ordered sequence of actions as defined in Eq.~\ref{eq:p_action}, where each action \(a_{t,k}\) is defined by a tool type, a target lane identifier, and tool-specific parameters. 
If all lanes are judged as correct, the Planner outputs an empty plan $\mathcal{P}_t=\emptyset$.

All planning decisions are constrained by externally defined and immutable mapping rules that encode lane-level specifications and safety constraints. 
These rules prohibit the creation of new lane lines, cross--lane-group modifications, and large-scale structural changes.
Every generated plan must satisfy both schema validity and rule compliance before execution; plans that fail validation are rejected and treated as empty plans.

The Planner maintains a lightweight memory of past refinement actions and their outcomes, which is used to filter redundant or conflicting decisions. 
If no valid plan can be constructed under the imposed constraints and current map state, the Planner outputs an empty plan and terminates refinement. 
This conservative fallback mechanism prevents cascading errors and reduces the risk of degrading the map when corrective actions are uncertain or inadmissible.

\subsection{Worker Agent: Deterministic Tool-Based Editing}
\label{sec:worker_agents}
MapAgent operates under a bounded refinement budget and a closed, specification-verified tool set. The Worker tools are deterministic lane-line editors that support only local, auditable modifications. They cannot create new lane instances or perform cross-lane-group/non-local topology modifications. Deletion is allowed only for lanes diagnosed as redundant or spurious and is accepted only after passing the feasibility gate $\Omega$. Therefore, the Workers are designed to avoid unsafe global topology changes while enabling local, auditable correction.

The Worker Agent is responsible for carrying out the refinement plans generated by the Planner Agent. 
MapAgent exposes a fixed and limited set of Worker tools, each designed for a specific type of
lane-level refinement:
(i) a \emph{Category Worker} for semantic label correction,
(ii) a \emph{Delete Worker} for removing redundant or spurious lane lines,
(iii) a \emph{Smooth Worker} for local geometric smoothing, and
(iv) a \emph{Regenerate Worker} for local geometry repair using a trained model (SAM3)~\cite{carion2025sam}.
Among the available tools, the Regenerate Worker is the only one that leverages a learned model.
It is designed for local geometry repair within the spatial support of an existing lane line, and its behavior is bounded by Planner-specified constraints, targeting local inconsistencies such as broken segments or misaligned geometry. Refinement actions are executed sequentially according to the order specified in the plan
\(\mathcal{P}_t\).


\section{Experiments}
\label{sec:exp}

In this section, we describe the experimental setup, dataset, and evaluation protocol for MapAgent, followed by quantitative results and analysis. We also include ablations and case studies to better understand the refinement behavior under challenging scenes.

\subsection{Experimental Settings}

\textbf{Dataset Construction.}
Following the DuLD dataset construction protocol in DuMapNet~\cite{xia2024dumapnet}, we build large-scale lane-level vector maps from the Baidu Map Database by rendering a high-quality offline BEV image $I$ via multi-trajectory aggregation of camera--LiDAR fusion signals and assembling the vectorized ground truth $V^{\mathrm{GT}}$ from lane-related instances and attributes within each region (grouped by lane-group IDs, transformed into the local BEV coordinate system, and filtered with basic consistency operations such as removing invalid geometries and normalizing attributes). On top of this base dataset, we further curate a challenging hard subset using a backbone-agnostic difficulty criterion (e.g., high junction complexity and occlusion proxies derived from map topology / scene metadata), and use the same subset consistently across all backbones to ensure fairness. The training split contains 3{,}712 BEV images with 59{,}434 ground-truth lane instances; DuMapNet and GeMap produce 59{,}928 and 50{,}263 predicted lane instances, respectively. The test split contains 656 images with 10{,}254 ground-truth lanes; DuMapNet and GeMap produce 10{,}340 and 8{,}734 predicted lane instances, respectively.

\textbf{Metrics.}
We measure lane-level correction quality by matching each predicted lane to the most appropriate ground-truth lane and then computing metrics on the resulting assignments. Accuracy, Precision, Recall, and F1 are defined on fully correct lanes: a prediction is counted as a true positive only if it can be matched to a ground-truth lane and is correct as a whole, i.e., it satisfies the matching criterion and has the correct lane category. Predictions without a valid match are treated as false positives, and unmatched ground-truth lanes are treated as false negatives. In contrast, BBox IoU and Mask IoU quantify geometric overlap, and Cls Acc measures category correctness, all computed only over matched lane pairs.

\textbf{Backbones and Protocol.}
We evaluate MapAgent as a post-hoc refinement module on top of two representative BEV vectorization backbones, GeMap and DuMapNet.
Unless otherwise noted, backbone predictors are frozen and MapAgent is applied without retraining.
For the VLM-based Judge Agent, we compare different base models, including Qwen3-VL-Instruct (8B), Qwen3-VL-Thinking (8B), and InternVL-3.5-8B,
under an identical prompting and inference protocol (same prompt template, decoding strategy, and refinement budget).
All hyperparameters related to refinement (e.g., maximum number of tool calls and termination criteria) are fixed across experiments.

\begin{table*}[t]
\centering
\caption{Impact of VLM Judges and Fine-Tuning on Lane Quality Inspection.}
\label{tab:vlm_judge_comparison}
\resizebox{\textwidth}{!}{
\begin{tabular}{lcccccc}
\toprule
\multirow{2}{*}{\textbf{Judge Model}} 
& \multirow{2}{*}{\textbf{Accuracy (\%)}} 
& \multicolumn{5}{c}{\textbf{Precision / Recall (\%)}} \\
\cmidrule(lr){3-7}
& & No Error & Extra Lane Line & Category Error & Geometry Error & Structure Error \\

\midrule
InternVL-3.5-8B (SFT) 
& 58.23 
& 65.00 / 82.80 
& 80.00 / 49.38 
& 66.67 / 54.05 
& 57.14 / 40.82 
& 31.87 / 55.43 \\
\midrule
Qwen3-VL-8B (SFT)
& 70.16
& 87.50 / 89.17
& 93.33 / 86.42
& 88.24 / 67.57
& 83.33 / 51.02
& 18.18 / 32.61 \\
\midrule
Qwen3-VL-8B-Thinking (SFT) 
& 83.55
& 84.39 / 92.99
& 91.67 / 81.48
& 88.04 / 72.97
& 81.25 / 79.59 
& \textcolor{MapAgentRed}{70.43} / \textcolor{MapAgentRed}{88.04} \\
\midrule
\cellcolor{MapAgentBG}{Qwen3-VL-8B-Thinking (GRPO) }
& \cellcolor{MapAgentBG}\textcolor{MapAgentRed}{86.01}
& \cellcolor{MapAgentBG}\textcolor{MapAgentRed}{92.31} / \textcolor{MapAgentRed}{94.90}
& \cellcolor{MapAgentBG}\textcolor{MapAgentRed}{96.15} / \textcolor{MapAgentRed}{85.80} 
& \cellcolor{MapAgentBG}\textcolor{MapAgentRed}{93.33} / \textcolor{MapAgentRed}{81.08} 
& \cellcolor{MapAgentBG}\textcolor{MapAgentRed}{87.10} / \textcolor{MapAgentRed}{82.65} 
& \cellcolor{MapAgentBG} 66.67 / 82.61 \\
\bottomrule
\end{tabular}
}
\end{table*}

\begin{table*}[t]
\centering
\caption{Compact ablation of MapAgent under unified evaluation metrics.}
\label{tab:mapagent_ablation_compact_v6}
\resizebox{\linewidth}{!}{
\begin{tabular}{l c c c c c c c}
\toprule
Variant
& Accuracy $\uparrow$
& Precision $\uparrow$
& Recall $\uparrow$
& F1-score $\uparrow$
& BBox IoU $\uparrow$
& Mask IoU $\uparrow$
& Cls Acc $\uparrow$ \\
\midrule

Base Predictor (w/o MapAgent)
& 52.5 & 71.7 & 66.5 & 68.9 & 70.4 & 35.0 & 90.0 \\

\midrule
\textbf{w/o Reason} (Judge predicts error type only)
& 58.4 & 77.1 & 70.9 & 73.7 & 71.2 & 35.5 & 94.8 \\

\midrule
\textbf{Max Correction Rounds} ($T$)
& \multicolumn{7}{l}{} \\
\quad $T{=}1$
& 58.3 & 76.9 & 70.8 & 73.6 & 71.2 & 35.6 & 94.5 \\
\quad $T{=}2$
& 60.3 & 78.7 & 72.2 & 75.2 & 71.7 & 35.7 & 97.5 \\
\rowcolor{MapAgentBG}
\textbf{\quad $T{=}3$}
& \textcolor{MapAgentRed}{62.6}
& \textcolor{MapAgentRed}{80.7}
& \textcolor{MapAgentRed}{73.9}
& \textcolor{MapAgentRed}{77.0}
& \textcolor{MapAgentRed}{71.8}
& \textcolor{MapAgentRed}{36.0}
& \textcolor{MapAgentRed}{98.0} \\

\bottomrule
\end{tabular}
}
\end{table*}

\subsection{Implementation Details}
\label{ssec:implementation}

\textit{Hardware and Software.}
All experiments were run on a single server equipped with $8\times$ NVIDIA A800 80GB GPUs.
The software stack is PyTorch~2.6.0 with CUDA~12.4.

\textit{VLM, SFT and GRPO.} The Judge Agent module is based on the model \texttt{Qwen3-VL-8B-Thinking} (approximately 8B parameters).

The model is initialized from a publicly available pretrained checkpoint and further adapted using parameter-efficient fine-tuning.
Specifically, we first apply supervised fine-tuning (SFT) with LoRA using a learning rate of $1\mathrm{e}{-4}$, batch size of $32$, LoRA rank $8$, and $2$ epochs.
On top of the SFT checkpoint, we perform a lightweight GRPO stage to better align Judge outputs with downstream map-refinement objectives.
GRPO updates only the LoRA parameters with learning rate $1\mathrm{e}{-5}$, rollout batch size $16$, GRPO clip $\epsilon=0.2$, KL coefficient $0.1$, and $3$ epochs per update.

\textit{Runtime Measurements.}
After $100$ warm-up tiles, we measure runtime over $1{,}000$ randomly sampled validation tiles (including data I/O).
The full MapAgent pipeline achieves a mean latency of $420$\,ms/tile, median $380$\,ms, p95 $920$\,ms, and p99 $1.6$\,s. Module-level averages are $230$\,ms per tile for the Judge Agent (p95 $540$\,ms) and $140$\,ms/tile for the Worker (SAM + regenerate, p95 $600$\,ms). Peak GPU memory usage is approximately $19$\,GB per A800.
MapAgent is triggered on about $30\%$ of tiles in the test set.

\subsection{Main Results}

Table~\ref{tab:vlm_judge_comparison} details the performance of VLM-based Judges on lane-quality inspection. 
SFT yields 58.23\% for InternVL-3.5-8B and 70.16\% for Qwen3-VL-8B, while subsequent GRPO alignment yields further targeted gains. 
For instance, GRPO improves the overall accuracy of Qwen3-VL-8B-Thinking from 83.55\% to 86.01\% and improves most class-wise precision/recall values. 
However, the Structure Error category shows a slight drop, suggesting a trade-off between overall decision accuracy and this minority/harder error type.

The comparison also shows that reasoning-oriented VLMs are more suitable for priority-based quality inspection. 
Qwen3-VL-8B-Thinking substantially outperforms the non-thinking Qwen3-VL-8B under SFT, indicating that explicit reasoning helps the Judge distinguish visually similar error types and follow the predefined short-circuit order. 
After GRPO, the model further improves on \texttt{no\_error}, \texttt{extra\_lane\_line}, \texttt{category\_error}, and \texttt{geometry\_error}, suggesting that reward-based alignment makes the Judge outputs more consistent with executable downstream refinement.

Table~\ref{tab:mapagent_unified} evaluates MapAgent as a frozen post-hoc refinement layer for GeMap and DuMapNet. 
MapAgent consistently improves lane-level correctness, scaling with Judge capability. 
On GeMap, Qwen3-VL-Thinking lifts Accuracy from 52.8 to 61.3 and F1 from 69.1 to 76.0. 
DuMapNet sees even larger gains (Accuracy 52.2$\to$63.9, F1 68.6$\to$78.0). 
Stable improvements across various models suggest MapAgent benefits systematically from stronger visual--language judgments rather than specific architectures.

Notably, gains are concentrated in Precision, Recall, and classification correctness, while geometric metrics (e.g., IoU) remain stable. 
This aligns with MapAgent's role as a specification-aware editor: it corrects spurious lanes, category mismatches, and match failures caused by local geometry/category errors rather than aggressively altering lane geometry. 
The pronounced improvements on DuMapNet further indicate the refinement loop is particularly effective at rectifying errors in initial draft maps.

\subsection{Ablation Study}

To better understand which parts of MapAgent drive the gains, we run compact ablations under the same unified evaluation protocol as Table~\ref{tab:mapagent_unified}. For readability, Table~\ref{tab:mapagent_ablation_compact_v6} reports the averaged results over GeMap and DuMapNet, and we only show the final map quality after post-hoc refinement. These ablations focus on two key factors: whether the Judge provides structured evidence beyond error labels, and how much benefit can be obtained from iterative correction.

We first examine what happens when the Judge is reduced to a pure error-type classifier. When we remove the explicit reasoning/evidence generation and let the Judge output the error type only (w/o Reason), the system still improves noticeably over the frozen base predictors, moving Accuracy from 52.5 to 58.4 and F1 from 68.9 to 73.7, with Cls Acc rising from 90.0 to 94.8. However, the gap to the full MapAgent remains clear: with the complete Judge--Planner--Worker loop, F1 reaches 77.0 and Cls Acc reaches 98.0. This difference suggests that, beyond recognizing \emph{what} is wrong, the Judge's structured reasoning is important for producing actionable, localized evidence that the Planner can reliably translate into safe, tool-grounded edits. In other words, reasoning improves not only interpretability, but also the executability of subsequent correction.

We then vary the maximum correction budget $T$ in the bounded retry loop. A single round already captures a large portion of the benefit ($T{=}1$: 58.3 Accuracy and 73.6 F1), indicating many errors can be fixed with one pass of diagnosis and execution. Allowing a second iteration brings a further jump ($T{=}2$: 60.3 Accuracy and 75.2 F1), and a third round continues to help, though with diminishing returns ($T{=}3$: 62.6 Accuracy and 77.0 F1). This trend shows that iterative refinement is useful, but most easy-to-correct cases are resolved in the first few rounds. Across these settings, BBox/Mask IoU change only slightly (e.g., 70.4/35.0 at baseline to 71.8/36.0 at $T{=}3$), which aligns with MapAgent's conservative design: most gains come from resolving discrete false positives/negatives and type mistakes rather than aggressively deforming lane geometry or altering topology.

\begin{table*}[t]
\centering
\caption{Unified comparison of MapAgent as a post-hoc editing module across
different map prediction backbones and VLM bases.
MapAgent is applied to fixed backbone predictions without retraining.
Accuracy, Precision, Recall, F1-score, bounding box IoU (BBox IoU), mask IoU,
and classification accuracy (Cls Acc) are reported (\%).}
\label{tab:mapagent_unified}
\resizebox{\linewidth}{!}{
\begin{tabular}{l l c c c c c c c}
\toprule
Map Backbone & MapAgent Base
& Accuracy & Precision & Recall & F1-score
& BBox IoU & Mask IoU & Cls Acc \\
\midrule
\multirow{4}{*}{GeMap}
& Original Prediction
& 52.8 & 75.1 & 64.0 & 69.1 & 69.3 & 32.9 & 91.9 \\
& InternVL-3.5-8B
& 54.9 & 77.1 & 65.6 & 70.8 & 69.7 & 33.3 & 96.5 \\
& Qwen3-VL-Instruct (8B) 
& 56.5 & 78.6 & 66.8 & 72.2 & 69.8 & 33.5 & 96.8 \\

& Qwen3-VL-Thinking (8B)
& \cellcolor{MapAgentBG}\textcolor{MapAgentRed}{61.3}
& \cellcolor{MapAgentBG}\textcolor{MapAgentRed}{82.9}
& \cellcolor{MapAgentBG}\textcolor{MapAgentRed}{70.1}
& \cellcolor{MapAgentBG}\textcolor{MapAgentRed}{76.0}
& \cellcolor{MapAgentBG}\textcolor{MapAgentRed}{70.7}
& \cellcolor{MapAgentBG}\textcolor{MapAgentRed}{34.2}
& \cellcolor{MapAgentBG}\textcolor{MapAgentRed}{98.1} \\
\midrule
\multirow{4}{*}{DuMapNet}
& Original Prediction
& 52.2 & 68.3 & 68.9 & 68.6 & 71.4 & 37.1 & 88.0 \\
& InternVL-3.5-8B
& 55.0 & 70.9 & 71.1 & 71.0 & 71.9 & 37.4 & 94.6 \\
& Qwen3-VL-Instruct (8B)
& 57.1 & 72.7 & 72.7 & 72.7 & 72.1 & 37.4 & 95.3 \\
& Qwen3-VL-Thinking (8B)
& \cellcolor{MapAgentBG}\textcolor{MapAgentRed}{63.9}
& \cellcolor{MapAgentBG}\textcolor{MapAgentRed}{78.4}
& \cellcolor{MapAgentBG}\textcolor{MapAgentRed}{77.6}
& \cellcolor{MapAgentBG}\textcolor{MapAgentRed}{78.0}
& \cellcolor{MapAgentBG}\textcolor{MapAgentRed}{72.8}
& \cellcolor{MapAgentBG}\textcolor{MapAgentRed}{37.7}
& \cellcolor{MapAgentBG}\textcolor{MapAgentRed}{97.8} \\
\bottomrule
\end{tabular}
}
\end{table*}

\subsection{Case Study}

\begin{figure*}[t]
    \centering
    \begin{minipage}{0.19\linewidth}
        \centering
        \includegraphics[width=\linewidth]{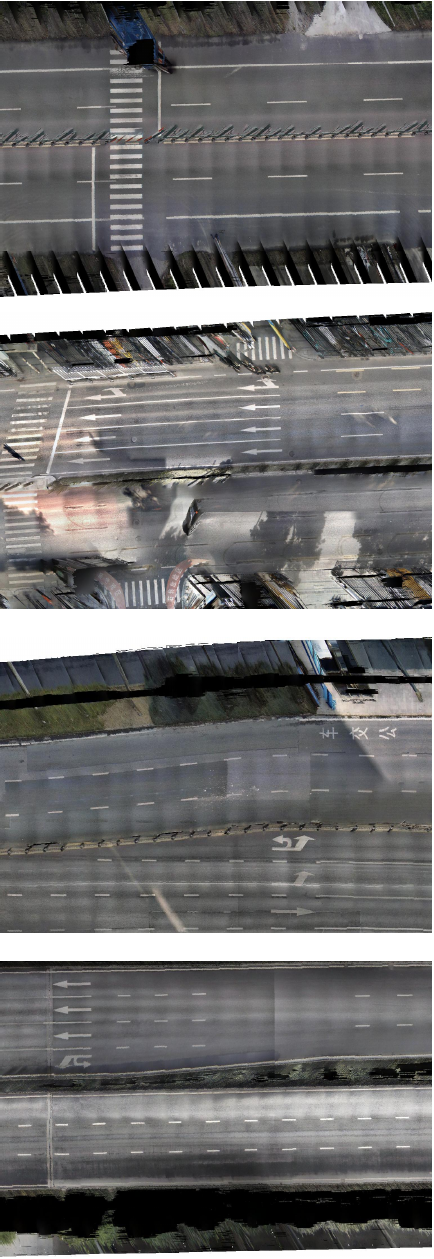}
        \subcaption{Origin}
        \label{fig:case_origin}
    \end{minipage}
    \hfill
    \begin{minipage}{0.19\linewidth}
        \centering
        \includegraphics[width=\linewidth]{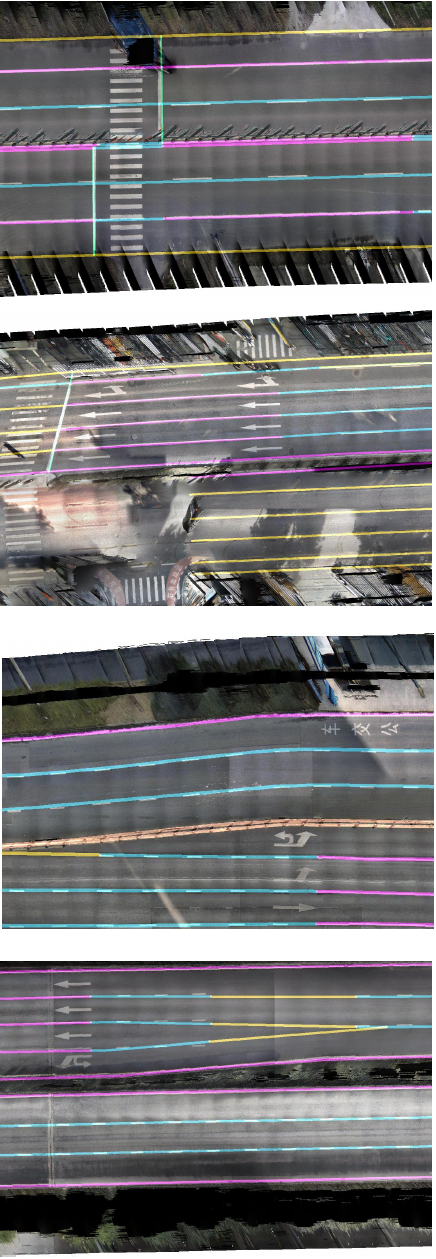}
        \subcaption{Ground Truth}
        \label{fig:case_gt}
    \end{minipage}
    \hfill
    \begin{minipage}{0.19\linewidth}
        \centering
        \includegraphics[width=\linewidth]{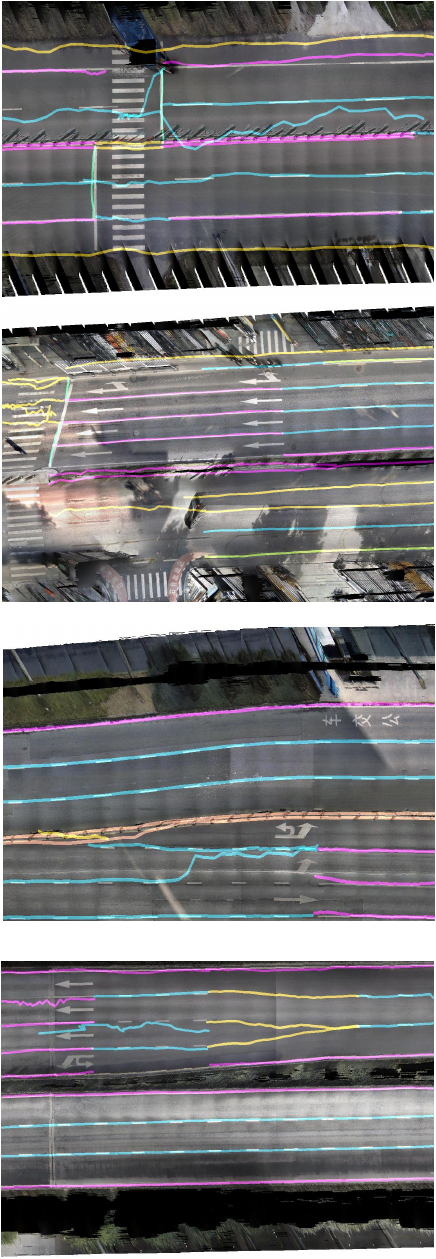}
        \subcaption{GeMap}
        \label{fig:case_gemap}
    \end{minipage}
    \hfill
    \begin{minipage}{0.19\linewidth}
        \centering
        \includegraphics[width=\linewidth]{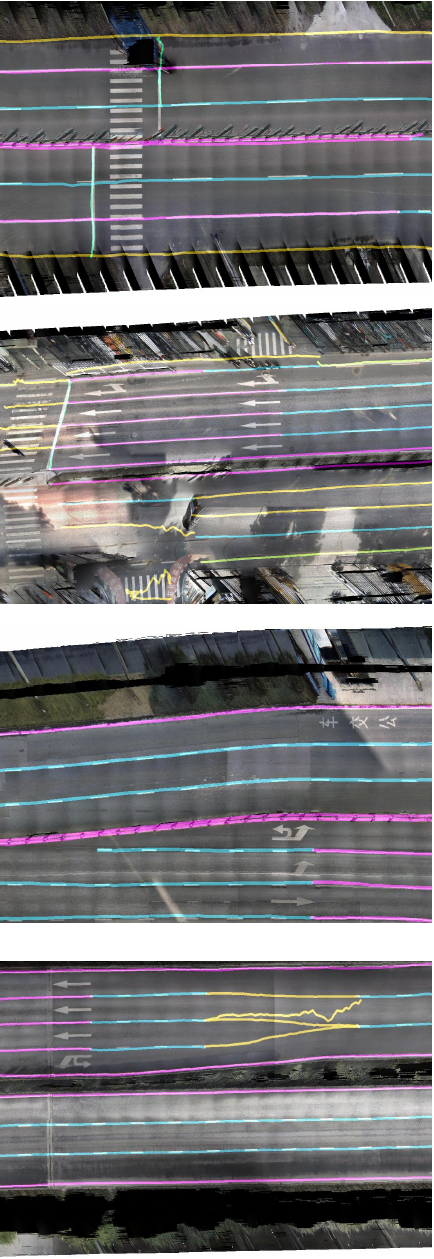}
        \subcaption{DuMapNet}
        \label{fig:case_dumap}
    \end{minipage}
    \hfill
    \begin{minipage}{0.19\linewidth}
        \centering
        \includegraphics[width=\linewidth]{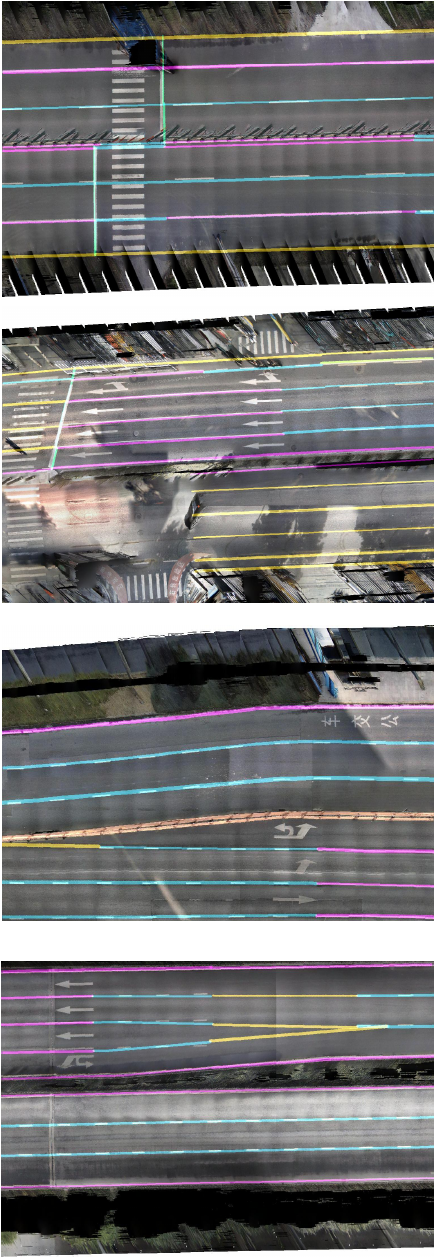}
        \subcaption{Corrected (Ours)}
        \label{fig:case_corrected}
    \end{minipage}
    \caption{
    Case study of lane-level map refinement under challenging scenarios.
    We compare the original input, ground-truth annotations, predictions from GeMap and DuMapNet,
    and the refined results produced by our correction framework.
    Best viewed in color and zoomed in.
    }
    \label{fig:case_study}
\end{figure*}

Figure~\ref{fig:case_study} shows several examples from the hard subset. As scenes become more challenging—e.g., with faint, missing, or occluded lane markings—both GeMap and DuMapNet often produce messy predictions, including spurious lanes, fragmented segments, and inconsistent topology. This failure mode is not specific to a particular backbone, but reflects a general limitation of feed-forward map prediction under weak or ambiguous visual evidence.

MapAgent effectively refines these noisy predictions into cleaner and more structured maps. The corrected results contain fewer spurious segments and exhibit more consistent global topology. In practice, the refinement focuses on obvious structural issues—removing hallucinated lanes, suppressing fragments, and restoring coherent layouts—while keeping geometry conservative. This qualitative trend matches the quantitative gains in lane-level correctness and demonstrates the method’s scalability.

\section{Discussion}
\label{sec:discussion}

MapAgent bridges the gap between BEV vectorization backbones and production-grade, specification-compliant lane maps. Integrated into Baidu Maps for city-scale lane-level map generation and updating, it supports 360+ cities and elevates automation to over $95\%$. Here, the automation rate is defined as the ratio of lane-level mileage completed fully automatically without human intervention to the total lane-level mileage over a fixed production period. Instead of one-shot prediction followed by heavy human post-editing, MapAgent adds a constrained Judge--Planner--Worker refinement layer: violations are diagnosed under factorized constraints, corrected via a closed set of deterministic tools, and re-validated within a bounded budget. By triggering refinement only on hard tiles, it preserves throughput while reducing manual correction workload with predictable latency.

Despite these advances, several challenges remain worth exploring to further enhance autonomy. 
First, extreme visual ambiguity can make \emph{lane addition} and \emph{topology modification} under-determined from visual evidence, requiring stronger priors and principled uncertainty handling to avoid unsafe edits. In the current production setting, MapAgent therefore prioritizes the primary manual post-editing categories in our pipeline that can be handled safely with deterministic, local tools while deferring lane addition and non-local topology modification to future work. 
Second, to meet large-scale engineering requirements, MapAgent currently refines outputs from an existing backbone; an important direction is a unified agentic framework capable of autonomously scheduling and orchestrating different specialized perception backbones (e.g., specialized backbones for junctions, occlusions, or rare topologies) and integrating their complementary predictions into a single constrained generation-and-updating pipeline, thereby maximizing flexibility and throughput.

\section{Related Work}
\label{sec:rw}

\subsection{Map Construction}
High-definition map construction has evolved from multi-stage segmentation pipelines to end-to-end vectorized regression~\cite{li2022hdmapnet}. VectorMapNet~\cite{liu2023vectormapnet} and MapTR~\cite{liao2022maptr} advanced direct polyline decoding with autoregressive and permutation-invariant transformers, respectively, while recent systems further improve geometric consistency and structured modeling, e.g., GeMap~\cite{zhang2023gemap} and HiMap~\cite{zhou2024himap}. 
Beyond purely visual cues, a growing line of work leverages \emph{map priors} to stabilize matching and extend range, including Neural Map Prior~\cite{xiong2023neural}, P-MapNet~\cite{jiang2024p}, and PriorMapNet~\cite{wang2024priormapnet}, which incorporate SD maps or historical/outdated priors as additional conditioning signals. More recently, SDTagNet~\cite{immel2025sdtagnet} exploits text-annotated SD maps to enhance far-range online HD map construction. 
In parallel, mixture-of-experts and interaction-based designs (e.g., MapExpert~\cite{zhang2025mapexpert} and InteractionMap~\cite{wu2025interactionmap}) improve long-tail element modeling via expert routing and structured temporal--spatial interactions. 
Concurrently, VLM-based efforts such as MapGPT~\cite{zhang2024mapgpt} and MAPLM~\cite{cao2024maplm} suggest that multimodal reasoning may help interpret maps and traffic scenes.
However, these methods largely remain \emph{monolithic predictors} that optimize geometric/semantic metrics and treat cartographic standards and traffic regulations as implicit supervision, leaving specification violations to manual post-editing. 
MapAgent addresses this gap by serving as a specification-aware refinement layer on top of strong BEV backbones.

\subsection{Agentic System}
LLM/VLM-based agents have rapidly advanced in tool use and self-correction.
ReAct~\cite{yao2022react} interleaves reasoning and acting for interactive problem solving, Toolformer~\cite{schick2023toolformer} learns to decide when/how to call tools, and Reflexion~\cite{shinn2023reflexion} improves agents via feedback-driven memory.
Recent work further strengthens intrinsic self-correction via multi-turn RL (e.g., SCoRe~\cite{kumar2024training}).
Complementary lines study modular tool routing and grounded execution (e.g., MRKL~\cite{karpas2022mrkl}, SayCan~\cite{ahn2022can}, ReWOO~\cite{xu2023rewoo}, Voyager~\cite{wang2023voyager}). 
Meanwhile, general-purpose multimodal foundations (e.g., GPT-4~\cite{achiam2023gpt4}, Qwen-VL~\cite{bai2023qwenvl}, InternVL~\cite{chen2024internvl}, PaLM-E~\cite{driess2023palm}) provide strong visual grounding and verification capabilities.
While agentic workflows have been applied to driving planning (e.g., DriveLM~\cite{sima2024drivelm}), they have not been adapted for the strict safety constraints of map production. MapAgent pioneers this paradigm in mapping by combining VLM-based specification verification with deterministic tool execution, ensuring industrial-grade compliance.

\section{Conclusions}
\label{sec:concl}
In this paper, we present MapAgent, an industrial-grade agentic framework that bridges end-to-end vectorization and specification-compliant map production. Motivated by the limitations of purely data-driven backbones in complex scenes, we formulate map refinement as a bounded, verification-driven iterative refinement process and implement MapAgent as refinement-on-top-of-backbone. A lightweight Quality Agent selects hard tiles, and a bounded Judge--Planner--Worker loop refines them: a structured vision–language Judge provides grounded diagnosis, a constrained Planner generates tool-grounded edits, and deterministic Workers execute them safely to enforce geometric/topological validity and traffic standards. Experiments on large-scale real-world datasets show consistent gains over strong production baselines, especially in long-tail scenarios.
Furthermore, MapAgent has been successfully integrated into the production pipeline of Baidu Maps. Supporting lane-level map generation and updating for over $360$ cities nationwide, it has elevated the overall production automation to over $95\%$, proving that agent-based refinement is a viable and efficient paradigm for large-scale autonomous driving infrastructure. 
These results highlight the effectiveness of structured agentic refinement for improving map quality under challenging conditions. We believe this paradigm provides a practical and scalable direction for future large-scale mapping systems.

\section{Acknowledgments}
This work was supported by Beijing Natural Science Foundation (L231008, L243008), National Natural Science Foundation of China (52472449, 52402499), Independent Research Project of the State Key Laboratory of Intelligent Green Vehicle and Mobility, Tsinghua University (No. ZZ-PY-20250408), the Tsinghua University-Toyota Joint Center, and Tsinghua University--SAIC GM Wuling Joint Research Center.

\clearpage

\balance
\bibliographystyle{ACM-Reference-Format}
\bibliography{main}

\clearpage
\appendix
\counterwithin{figure}{section}
\counterwithin{table}{section}

\section{Feasibility Check and Specification Library}
\label{apsec:omega}

In MapAgent problem formulation, we implement feasibility as a boolean quality gate. Let $\Omega:\mathcal{V}\rightarrow\{0,1\}$ denote a QC function induced by a versioned library of
\emph{hard} (boolean) specification predicates $\mathcal{R}^{(v)}=\{r_k\}_{k=1}^{K}$,
derived from industry standards and internal cartographic specifications. We define
\begin{equation}
\begin{gathered}
    \Omega(V) \triangleq \mathtt{GeoValid}(V) \wedge \mathtt{TopoValid}(V) \wedge \mathtt{SpecValid}(V), \quad \\
    \mathtt{SpecValid}(V) \triangleq \bigwedge_{k=1}^{K} r_k(V), \quad
\end{gathered}
\end{equation}
where $\texttt{GeoValid}$ and $\texttt{TopoValid}$ are lightweight geometric / topological sanity checks
(e.g., no self-intersection, bounded curvature/length, lane-group consistency), and each $r_i(V)\in\{0,1\}$
encodes a non-negotiable cartographic / traffic constraint.

Given a deterministic tool transition $V_{t+1}=\mathcal{T}(V_t,a_t)$, each action must satisfy
$\mathtt{ActionValid}(a_t,V_t)$ (schema / parameter validity, bounded edit magnitude, lane-group scope),
and an updated state is accepted only if it passes the QC gate, i.e., $\Omega(V_{t+1})=1$.

\section{SAM3 Fine-tuning for Lane Detection}

\begin{figure}[h]
    \centering
    \includegraphics[width=\linewidth]{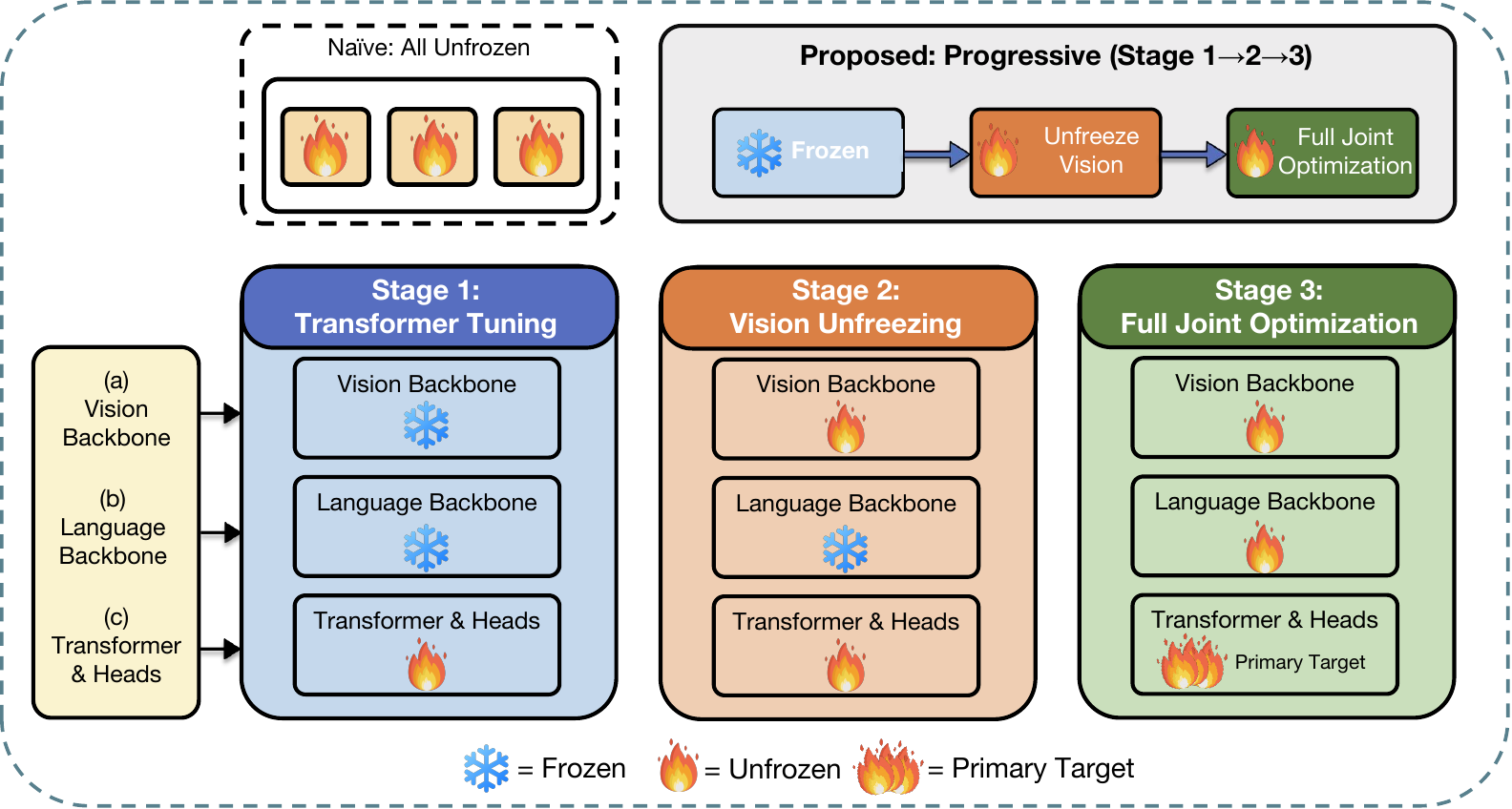}
    \caption{Architecture and fine-tuning strategy of SAM3 for lane detection.}
    \label{fig:sam3}
\end{figure}

This part describes the fine-tuning strategy of SAM3 for lane detection. Figure~\ref{fig:sam3} illustrates the overall architecture and the progressive fine-tuning design, where different stages gradually relax the optimization constraints on the pretrained components. 
We provide detailed training configurations and the rationale behind each stage, followed by a quantitative comparison in Table~\ref{tab:sam3_compare}. 
The results demonstrate that progressively unfreezing the backbones leads to consistent improvements across detection, segmentation, and classification metrics. Related training configurations and examples have been released at: {\color{blue}\url{https://github.com/eadst/KDD-2026-MapAgent}}.

\subsection{Training setup}
All stages are initialized from the same pretrained SAM3 checkpoint (sam3.pt) and trained on the lane dataset specified by \texttt{train.json} and \texttt{val.json}. We enable segmentation supervision throughout (\texttt{enable\_segmentation=True}) and load RLE masks during batching (\texttt{with\_seg\_masks=True}). Input images are resized to a square resolution of 1008 with a minimum scale of 480, followed by padding to the same size and normalization with mean and std both set to $(0.5,0.5,0.5)$. During training, we perturb input boxes with Gaussian noise (std 0.1, capped at 20 pixels) and filter samples with empty targets; we also cap the maximum number of lane instances per image to 30.

We optimize with AdamW and mixed precision enabled (bfloat16), using gradient clipping with max norm 1.0. The learning rate is scheduled by an inverse square root scheduler with 2000 warmup steps and timescale 10000. We use a base learning rate of $2\times 10^{-5}$ for transformer layers, and apply smaller learning rates for backbones when they are unfrozen: $6.25\times 10^{-6}$ for the vision backbone and $1.25\times 10^{-6}$ for the language backbone. Stage~1 updates only the transformer layers, Stage~2 additionally unfreezes the vision backbone with layer-wise decay 0.7, and Stage~3 further unfreezes the language backbone with the same conservative scaling. Weight decay is set to 0.05, while biases and LayerNorm parameters are excluded from decay. We train for 120 epochs with a fixed random seed of 123.

For matching and loss computation, we use a binary Hungarian matcher with focal loss $(\alpha=0.3, \gamma=2.0)$, augmented by an auxiliary one-to-many branch with weight 2.0, top-$k=4$, and threshold 0.4. The loss weights are set to 3.0 for $\mathcal{L}_{\text{bbox}}$ and 1.5 for $\mathcal{L}_{\text{giou}}$, 15.0 for classification and presence losses, and we strengthen mask supervision with $\mathcal{L}_{\text{mask}}$ weight 8.0 and $\mathcal{L}_{\text{dice}}$ weight 12.0. To better fit thin lane structures, mask loss is computed with point sampling (12544 points, oversample ratio 3, importance sampling ratio 0.75), focusing updates on uncertain boundary regions. All experiments are trained with distributed data parallel using NCCL on 8 GPUs per node, batch size 2 per GPU (global batch 16), and we keep all remaining settings fixed across stages to ensure fair comparison.

\subsection{Progressive Fine-tuning Strategies}
To adapt SAM3 to the lane detection task, we investigate three fine-tuning strategies that differ in how much of the pretrained model is allowed to adapt. Rather than treating fine-tuning as a single configuration choice, we gradually relax the optimization constraints to balance representation stability and task-specific adaptability.

\textbf{Stage 1.}
We begin with a conservative setting in which both the vision and language backbones are kept fixed, and only the transformer layers are optimized. In this configuration, the model primarily leverages pretrained visual representations while learning to reorganize cross-modal interactions for lane detection. This setting emphasizes stability and provides a reliable reference, revealing the extent to which SAM3 can model thin and elongated lane structures without modifying its backbone features.

\textbf{Stage 2.}
We then allow limited adaptation of the vision backbone while keeping the language backbone frozen. Specifically, the vision backbone is trained with a smaller learning rate and layer-wise decay, such that deeper layers receive progressively weaker updates. This design acknowledges that lane detection introduces domain-specific geometric patterns, including strong perspective effects and long-range continuity, which benefit from moderate refinement of visual features without disrupting low-level representations.

\begin{table*}[t]
\centering
\caption{Lane detection performance of SAM3 under the three progressive fine-tuning stages (Stage~1--3). Stage~1 updates only the transformer layers, Stage~2 additionally adapts the vision backbone with conservative optimization, and Stage~3 enables carefully scaled updates for all major components.}
\label{tab:sam3_compare}
\resizebox{\linewidth}{!}{
\begin{tabular}{l c c c c c c c}
\toprule
Model
& Accuracy (\%)
& Precision (\%)
& Recall (\%)
& F1-score (\%)
& BBox IoU (\%)
& Mask IoU (\%)
& Cls Acc (\%) \\
\midrule
SAM3 (Stage 1)
& 50.7 & 70.2 & 60.1 & 64.8 & 67.2 & 30.8 & 92.9 \\
SAM3 (Stage 2)
& 54.1 & 74.5 & 63.4 & 68.5 & 68.4 & 31.9 & 94.8 \\
SAM3 (Stage 3)
& \textbf{58.4} & \textbf{80.1} & \textbf{67.0} & \textbf{73.0} & \textbf{69.5} & \textbf{33.0} & \textbf{96.6} \\
\bottomrule
\end{tabular}
}
\end{table*}

\textbf{Stage 3.}
Finally, we consider a more flexible fine-tuning strategy in which all major components are updated with carefully scaled learning rates. The transformer remains the primary optimization target, while the vision and language backbones are adjusted more conservatively. This asymmetric optimization reflects their different roles in the task. The vision backbone focuses on refining geometric perception, whereas limited adaptation of the language backbone improves alignment between textual queries and fine-grained visual evidence in complex scenes. Despite its increased flexibility, this strategy remains well regularized and avoids excessive drift from pretrained semantics.

Overall, these strategies form a progressive fine-tuning spectrum that balances stability and adaptability. In practice, this design leads to stable optimization and provides insights into the contribution of different model components to lane detection performance.

\section{Chain-of-Thought Data Generation with GPT-5.2}
\label{app:cot_generation}
\begin{figure}[h]
    \centering
    \includegraphics[width=0.48\textwidth]{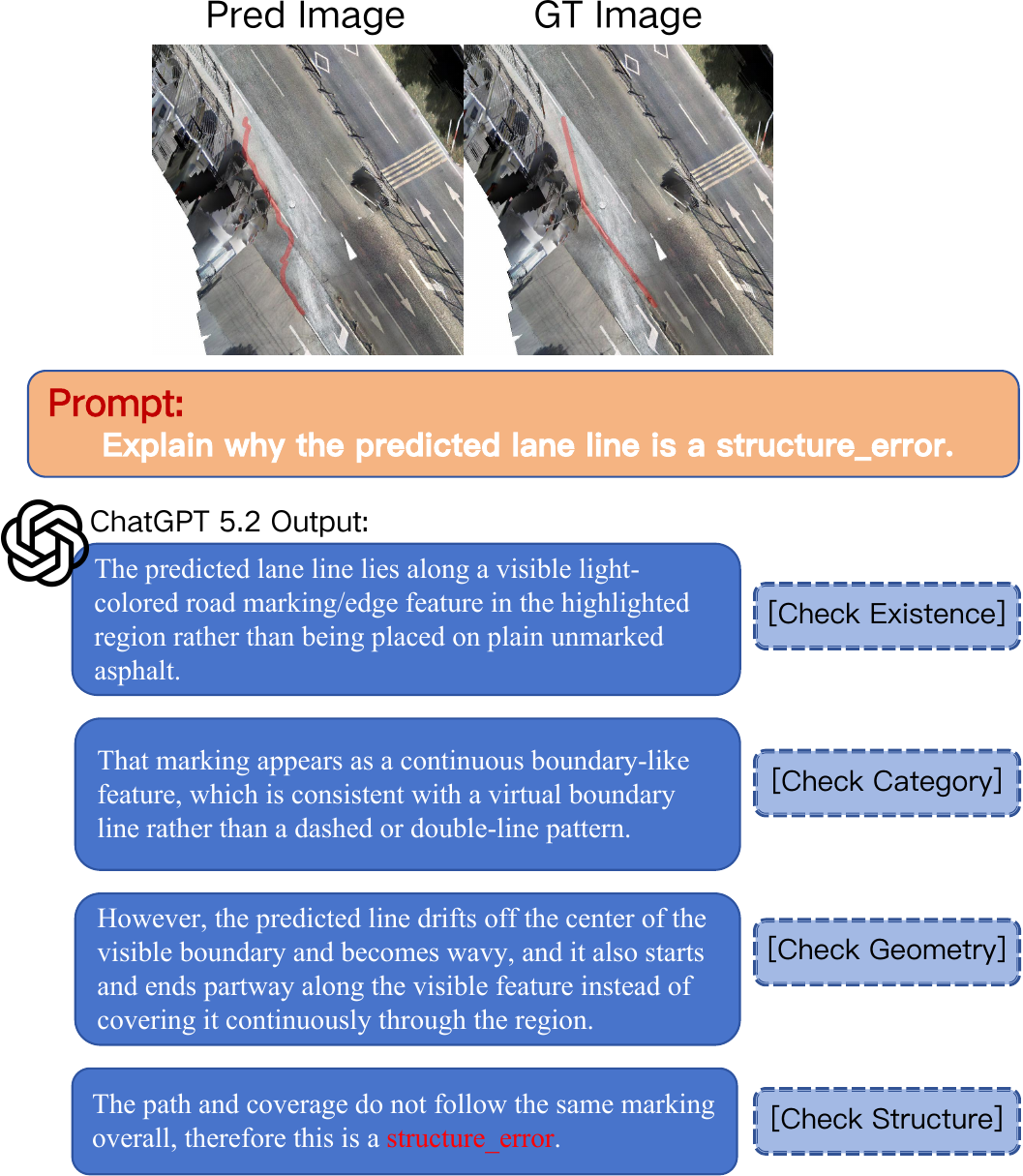}
    \caption{
    Example of priority-structured chain-of-thought generated by GPT~5.2 for Judge supervision.
    Given a predicted lane line (red overlay), the model produces a four-sentence explanation that follows a fixed
    elimination order: existence, category, geometry, and structure.
    Once a higher-priority condition is satisfied, the reasoning short-circuits, yielding a final
    \texttt{structure\_error} conclusion.
    For clarity, the prompt shown in the figure is a simplified illustrative version; the actual training prompt
    enforces the same reasoning structure and constraints.
    }
    \label{fig:cot_example}

\end{figure}
To supervise the Judge Agent with structured and priority-aware reasoning traces, we construct a chain-of-thought (CoT) dataset using GPT~5.2 as an offline data generator.
Rather than inferring labels, the model is tasked with producing concise visual rationales that \emph{justify a fixed oracle outcome}.
This design ensures that the generated CoT reflects the Judge’s diagnostic logic without introducing label ambiguity or decision leakage.

\subsection{Problem Setup}

Each training instance corresponds to a single predicted lane line produced by the frozen backbone model.
The input to GPT~5.2 includes:
(i) a \emph{Pred image} containing exactly one highlighted predicted lane line,
(ii) a \emph{GT image} used for supervision only,
and (iii) structured metadata, including the predicted category, a tight bounding box around the predicted line, and an oracle annotation specifying correctness and error type.

Importantly, the oracle annotation is treated as immutable.
GPT~5.2 is explicitly instructed not to question, revise, or infer the outcome, but only to generate a textual explanation that is consistent with the given oracle decision.

\subsection{Prompt Design Principles}
\label{app:prompt design}

A single unified system prompt is used to ensure consistency and controllability across all generated CoT samples.
The prompt enforces the following core principles.

\paragraph{Pred-only visual grounding.}
Although both Pred and GT images are available to the model to understand the scene, the generated explanation is constrained to rely \emph{only} on observable cues in the Pred image.
Explicit prohibitions prevent any reference to ground truth, labels, or cross-image comparison, ensuring that the resulting CoT is visually grounded and safe for deployment.

\paragraph{Priority-based short-circuit reasoning.}
The reasoning process follows a strict, ordered error taxonomy:
\begin{enumerate}
    \item \texttt{extra\_lane\_line}
    \item \texttt{category\_error}
    \item \texttt{geometry\_error}
    \item \texttt{structure\_error}
    \item \texttt{no\_error}
\end{enumerate}
Once a higher-priority error is identified, all lower-priority checks must be skipped and must not be mentioned.
This short-circuit structure mirrors the Judge Agent’s inference-time behavior and prevents mixed or internally inconsistent explanations.

\paragraph{Fixed-length structured rationale.}
Each CoT explanation is constrained to exactly four sentences.
The initial sentences rule out higher-priority error types using visible evidence, while the final sentence explicitly states the oracle error type as a conclusion.
This fixed structure simplifies downstream parsing and stabilizes supervised training.

\paragraph{Concrete and verifiable language.}
Abstract or hedging expressions are disallowed.
Instead, the model is required to describe concrete, image-verifiable phenomena such as early termination, over-extension, curvature deviation, gaps, misalignment, or partial coverage of visible markings.

\subsection{Prompt Instantiation}

Given the system prompt, each instance is instantiated with a lightweight task prompt that injects per-sample metadata, including the predicted category name, bounding box, and oracle error type.
Oracle-related fields are marked as \emph{do not question} and \emph{do not mention}, reinforcing that the model’s role is explanation rather than decision-making.

The output is constrained to a single JSON object of the form:
\[
\{\texttt{"reason"}: \text{string}\},
\]
where the value contains exactly four sentences.

\subsection{Qualitative Example}

Figure~\ref{fig:cot_example} illustrates a representative CoT example generated by GPT~5.2.
The explanation follows the prescribed priority order, first ruling out existence, category, and local geometry issues, and then short-circuiting at a \texttt{structure\_error} due to global path and coverage inconsistency.
The rationale is strictly grounded in observable cues from the Pred image and does not reference any ground-truth annotations.

\subsection{Resulting CoT Dataset}

The resulting CoT dataset consists of concise and deterministic rationales that:
(i) are strictly grounded in the Pred image,
(ii) align exactly with the oracle error labels,
and (iii) follow the same priority-structured decision logic as the Judge Agent.

During training, these CoT traces are used only as supervised reasoning targets.
They are not exposed to downstream Planner or Worker modules at inference time, preserving a clean separation between interpretable diagnosis and executable system actions.

\section{Bad Cases}
\label{app:bad_cases}

\begin{figure*}[t]
    \centering
    \includegraphics[width=\linewidth]{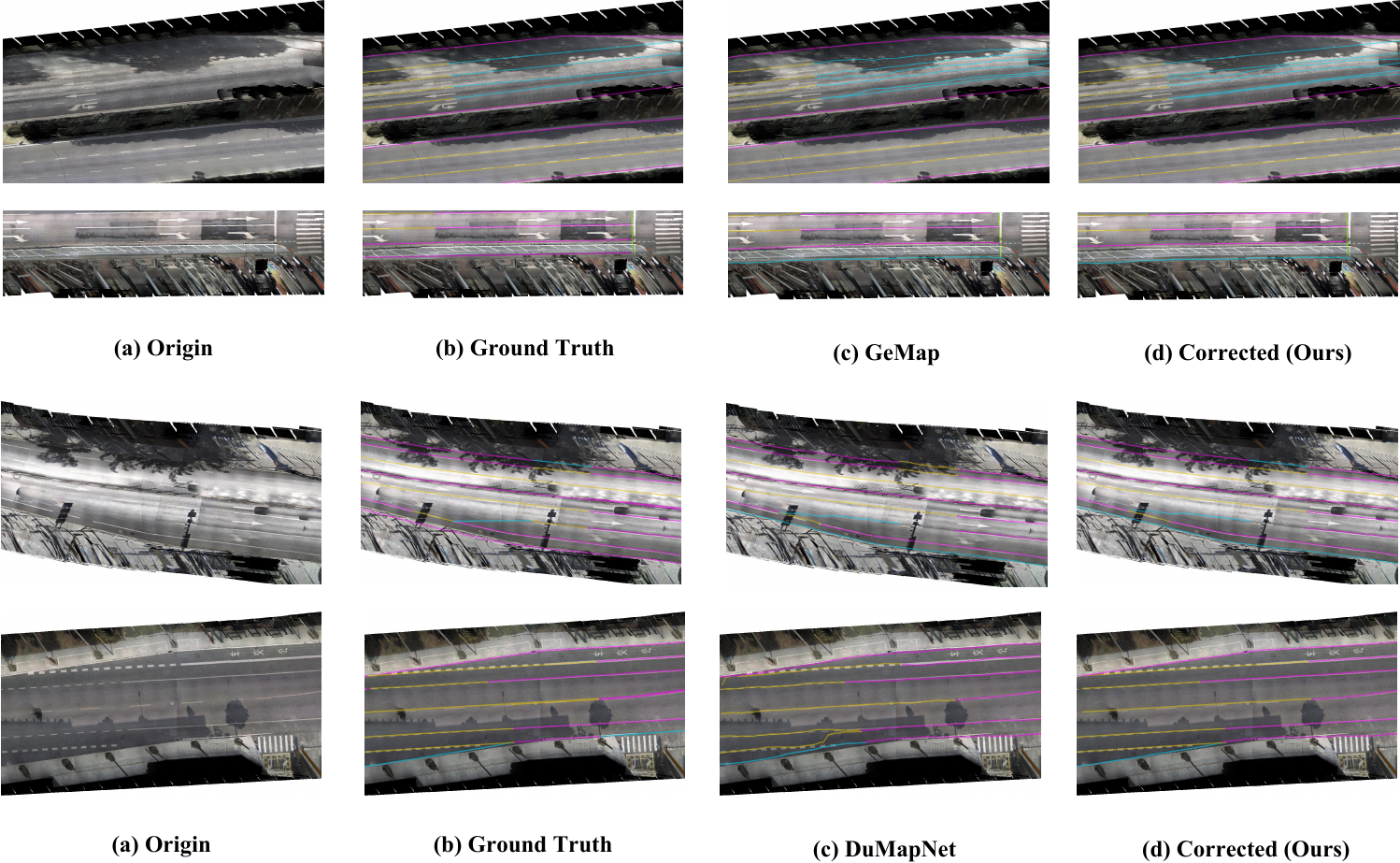}
    \caption{
    Representative bad cases under challenging road conditions. 
    The examples illustrate scenarios with weak visual evidence, ambiguous lane topology, and large deviations in the initial predictions, where refinement remains difficult. 
    Such cases reveal the practical limits of conservative correction and suggest directions for improving robustness in future work.
    }
    \label{fig:badcase}
\end{figure*}

Although MapAgent achieves strong overall performance, some challenging cases remain difficult. 
Figure~\ref{fig:badcase} shows representative failure cases where the scene contains weak lane evidence, severe shadows, local occlusions, or highly ambiguous topology. 
In such situations, the backbone predictions may already deviate substantially from the underlying lane layout, making reliable refinement more difficult. 
Since MapAgent is designed to apply conservative and tool-grounded edits, it tends to prioritize precision and structural consistency over aggressive topology rewriting. 
As a result, when the visual evidence is extremely limited or the initial prediction is severely corrupted, certain missing, shifted, or structurally inconsistent lane segments may still remain unresolved. 
These cases highlight an inherent limitation of refinement-based pipelines: the final quality still depends on both the recoverability of visual cues and the quality of the initial backbone prediction.
\section{Training-Time Prompt for Judge Supervision}
\label{app:judge_training_prompt}

To train the Judge Agent to produce priority-consistent quality assessments, we supervise it with the CoT dataset described in Section~\ref{app:cot_generation}. 
The generated rationales are used as reference reasoning traces, while the oracle error type and category provide structured supervision targets.

The training prompt presents each predicted lane line as a red semi-transparent mask overlaid on the BEV image. 
The model is instructed to treat the mask as a prediction hypothesis and to base its judgment on observable road evidence, including painted markings, curbs, asphalt boundaries, lane continuity, and local topology. 
This design encourages the Judge to verify whether the predicted lane is visually and structurally supported, rather than relying on the mask appearance alone.

The prompt follows the same priority-constrained evaluation rule used by the Judge Agent.
Once a higher-priority error is confirmed, all lower-priority checks are skipped.
This short-circuit mechanism prevents mixed diagnoses and keeps the training target consistent with inference-time decision logic.
For virtual lines, the prompt allows predictions that are topologically justified even without visible painted markings, and treats them as \texttt{extra\_lane\_line} only when no such justification exists.

The model is required to output a reasoning trace enclosed within \texttt{<think>} tags, followed by one final error type and a concise evidence summary. 
The reasoning trace is supervised by the generated CoT rationales, whereas the final error type and evidence summary are supervised by oracle annotations and reference explanations.

\end{document}